\pdfoutput=1

\documentclass[11pt]{article}

\usepackage{emnlp2021}

\usepackage{times}
\usepackage{latexsym}

\usepackage[T1]{fontenc}

\usepackage[utf8]{inputenc}

\usepackage{microtype}

\usepackage{graphicx}
\usepackage{bm}
\usepackage{amssymb}
\usepackage{amsmath}
\usepackage{multirow}
\newcommand{\ie}{\emph{i.e.}}
\newcommand{\eg}{\emph{e.g.}}

\newcommand{\aka}{\emph{aka}}

\title{Natural Language Video Localization with Learnable Moment Proposals}

\author{Shaoning Xiao$^\dagger$, Long Chen$^\ddagger$\thanks{\ \ Long Chen is the corresponding author. The work started when Long Chen was at Zhejiang University.}, Jian Shao$^\dagger$, Yueting Zhuang$^\dagger$, and Jun Xiao$^\dagger$ \\
$^\dagger$DCD Lab, College of Computer Science, Zhejiang University \\ $^\ddagger$DVMM Lab, Columbia University \\
{\tt shaoningx@zju.edu.cn, zjuchenlong@gmail.com} \\
{\tt jshao@zju.edu.cn, yzhuang@zju.edu.cn, junx@zju.edu.cn}
}

\begin{document}
\maketitle
\begin{abstract}
Given an untrimmed video and a natural language query, Natural Language Video Localization (NLVL) aims to identify the video moment described by the query. To address this task, existing methods can be roughly grouped into two groups: 1) \emph{propose-and-rank} models first define a set of hand-designed moment candidates and then find out the best-matching one. 2) \emph{proposal-free} models directly predict two temporal boundaries of the referential moment from frames. Currently, almost all the propose-and-rank methods have inferior performance than proposal-free counterparts. In this paper, we argue that propose-and-rank approach is underestimated due to the predefined manners: 1) Hand-designed rules are hard to guarantee the complete coverage of targeted segments. 2) Densely sampled candidate moments cause redundant computation and degrade the performance of ranking process. To this end, we propose a novel model termed \textbf{LPNet} (Learnable Proposal Network for NLVL) with a fixed set of learnable moment proposals. The position and length of these proposals are dynamically adjusted during training process. Moreover, a boundary-aware loss has been proposed to leverage frame-level information and further improve the performance. Extensive ablations on two challenging NLVL benchmarks have demonstrated the effectiveness of LPNet over existing state-of-the-art methods\footnote{Source codes is available: \url{https://github.com/xiaoneil/LPNet/}}.




\end{abstract}

\section{Introduction}

\begin{figure}
  \centering
  \includegraphics[width=0.48\textwidth]{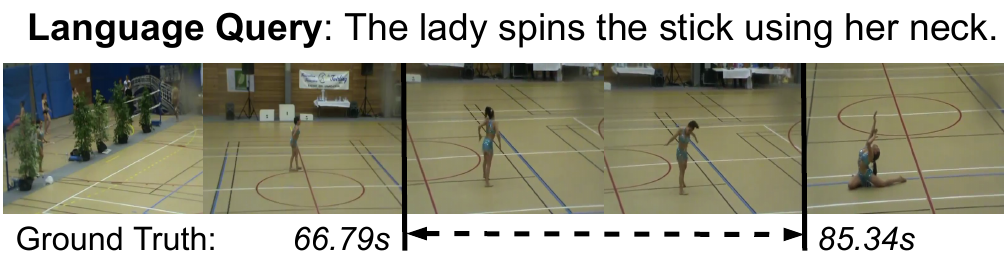}
  \caption{An illustrative example of NLVL. Given a video and a query, NLVL is to localize the video segment corresponding to the query with the start timestamp (66.79s) and the end timestamp (85.34s).}
  \label{figure1}
\end{figure}


Natural Language Video Localization (NLVL), \aka, video grounding or video moment localization, has got unprecedented attention in both CV and NLP communities~\cite{DBLP:conf/iccv/GaoSYN17,DBLP:conf/iccv/HendricksWSSDR17}. As shown in Figure~\ref{figure1}, NLVL aims to localize the video segment relevant to the query by locating the start and end timestamps in an untrimmed video. It is challenging since it needs to not only understand the video and the sentence content but also find out the precise temporal boundaries. Moreover, NLVL is helpful to numerous downstream video understanding tasks, \eg, content retrieval~\cite{DBLP:conf/eccv/ShaoXZHQL18}, relation detection~\cite{gao2021video}, and VQA~\cite{DBLP:conf/emnlp/LeiYBB18, ye2017video}.

Currently, state-of-the-art NLVL methods can be roughly grouped into two categories according to how the video segments are detected, namely \emph{propose-and-rank} and \emph{proposal-free} methods:

The idea of the \emph{propose-and-rank} approach~\cite{DBLP:conf/iccv/GaoSYN17,DBLP:conf/emnlp/HendricksWSSDR18,DBLP:conf/mm/LiuWN0CC18,ChenCMJC18,DBLP:conf/wacv/GeGCN19,DBLP:conf/aaai/Xu0PSSS19,DBLP:conf/cvpr/ZhangDWWD19} is intuitive,
which follows the same spirits of anchor-based object detectors, \eg, Faster R-CNN~\cite{DBLP:conf/nips/RenHGS15}.
This kind of methods firstly defines a series of manually-designed temporal bounding boxes as moment proposals. Then, they match each candidate with the sentence in a common feature space and compute matching scores for all the candidates. Thus the localization problem is reformulated into a ranking problem. However, these methods suffer from two inherent drawbacks due to the predefined manners: 1) Even though they elaborately design a series of hyperparameters (\eg, temporal scales and sample rates), these hand-designed rules are hard to guarantee the complete coverage of targeted video segments, and consequently tend to produce inaccurate boundaries. 2) A vast number of proposals are required to achieve high recall, which causes redundant computation and degrades the results of ranking process.

Another type of solution \emph{proposal-free} approach \cite{DBLP:conf/aaai/ChenJ19a,DBLP:conf/aaai/YuanM019,LuCTLX19,DBLP:conf/aaai/ChenLTXZTL20,DBLP:conf/acl/ZhangSJZ20} mitigates these defects. 
Instead of predefining a series of temporal proposals, they directly predict the start and the end boundaries or regress the locations of the query-related video segments. Benefit from such design, proposal-free methods get rid of placing superfluous temporal anchors, \ie, they are more computation-efficient. Furthermore, without fixing the position and length of the moment proposals, these methods are flexible to adapt to video segments with diverse lengths. Compared to propose-and-rank methods, there are two main limitations of proposal-free methods~\cite{DBLP:conf/aaai/XiaoCZJSYX21}: 1) They overlook the rich information between start and end boundaries because they are hard to model the segment-level interaction. 2) They always suffer from severe imbalance between the positive and negative training samples. 

Up to now, almost all the propose-and-rank methods have inferior performance. We argue that the performance of the propose-and-rank methods are underestimated due to current predefined designs. In this paper, we propose a novel propose-and-rank model with learnable moment proposals, termed \textbf{LPNet}. Without fixed dense proposals, only a sparse set of proposals are required to obtain decent performance.
In addition, there is no need to worry about the design of hyper-parameters because it is adaptable to targeted segments with diverse positions and lengths. Obviously, as a propose-and-rank method, LPNet also avoids the defects of the proposal-free approach. 

Specifically, LPNet places a fixed set of learnable temporal proposals represented by 2-d coordinates indicating the centers and lengths of video segments. These proposals are used to extract visual features of Moment of Interest (MoI). In order to model the relative relations among candidates, a module has been proposed to make the candidates interact with each other using the self-attention mechanism. Then an individual classifier is used to predict the matching score between these proposals and the sentence query. During the training process, the coordinates of the proposal with maximum score are adjusted by a dynamic adjustor at each iteration. After sufficient iterations, these learned moment proposals will statistically represent the prior distributions of ground-truth segments on the dataset. In addition, we empirically find that the propose-and-rank models always obtain sub-optimal results without frame-level supervision. A boundary-aware predictor has been proposed to regularize the model to utilize frame-level information, which further boosts the grounding performance. 

We demonstrate the effectiveness of our model on two challenging NLVL
benchmarks (Charades-STA, and ActivityNet Captions) by extensive 
ablative studies. Particularly, our model achieves new state-of-the-art 
performance over all datasets and evaluation metrics.

\begin{figure*}[!t]
    \centering
    \includegraphics[width=0.98\textwidth]{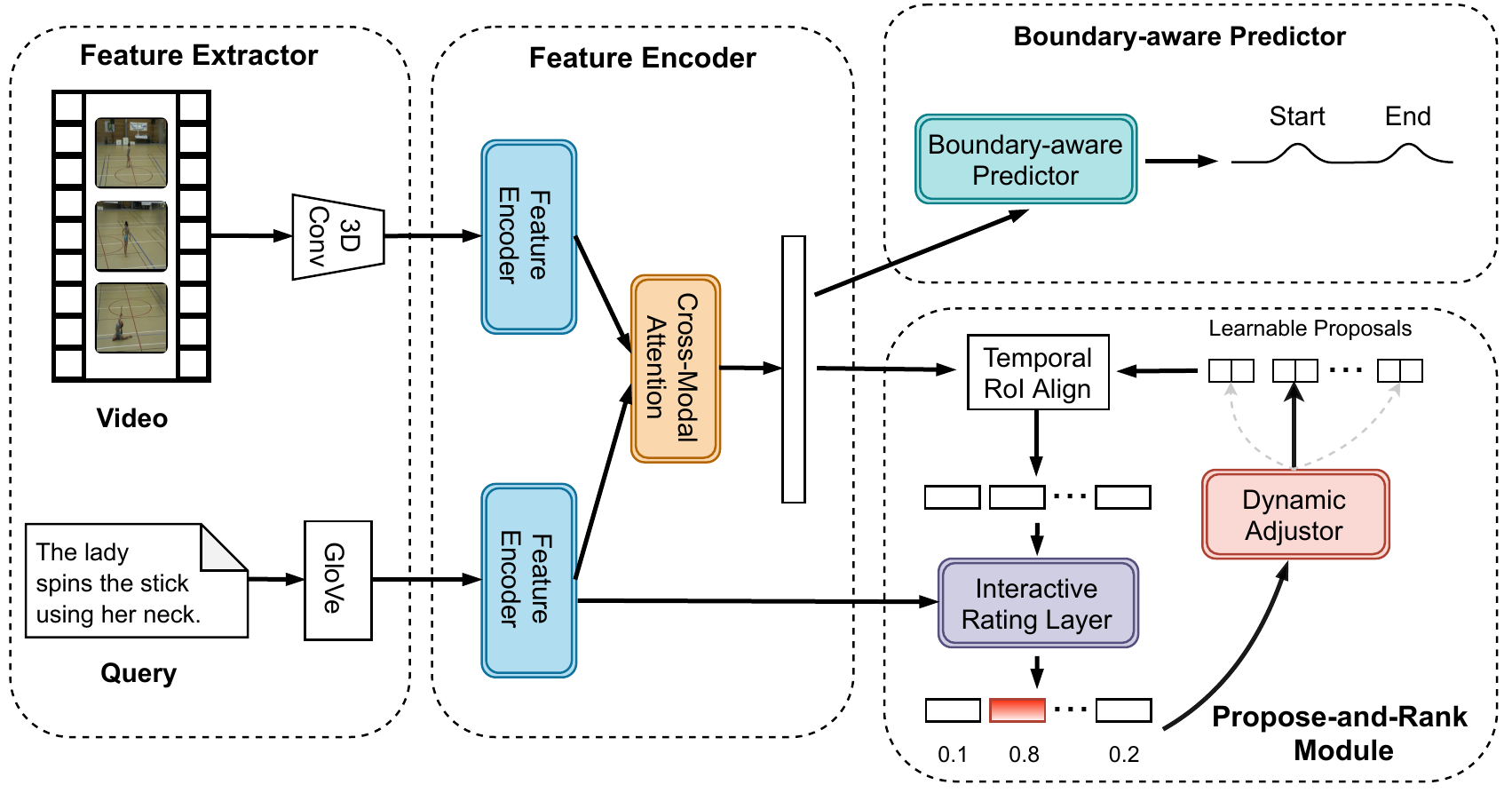}
    \caption{The architecture of LPNet for NLVL. Feature extractor transforms input video and language query into feature space. Feature encoder further refines video and language feature, and produces the multi-modal feature. A series of learnable proposal boxes are proposed which can be updated by dynamic adjustor during training. Interactive rating layer scores each candidate generated by proposal boxes and the candidate with highest score is the final prediction. Boundary-aware predictor takes multi-modal feature as input and predict the distribution of start and end timestamps, which is an auxiliary task to regularize the model to get better performance.
    }
    \label{figure2}
\end{figure*}

\section{Related Work}
\textbf{Natural Language Video Localization.}
NLVL task was first introduced in~\cite{DBLP:conf/iccv/HendricksWSSDR17,DBLP:conf/iccv/GaoSYN17}.
Current existing methods can be roughly grouped into two categories, namely \emph{propose-and-rank} and \emph{proposal-free} methods.

The propose-and-rank approaches~\cite{DBLP:conf/iccv/GaoSYN17,DBLP:conf/iccv/HendricksWSSDR17,DBLP:conf/emnlp/HendricksWSSDR18,DBLP:conf/mm/LiuWN0CC18,DBLP:conf/sigir/LiuWN0CC18,DBLP:conf/aaai/Xu0PSSS19,DBLP:conf/cvpr/ZhangDWWD19} solve the NLVL task by matching the predefined video moment proposals (\eg, in sliding window manner) with the language query and choose the best matching video segment as the final result. 
\citet{DBLP:conf/iccv/GaoSYN17} proposed a CTRL model. It takes video moments predefined through sliding windows as input and jointly models text
query and video clips, then outputs alignment scores and action
boundary regression results for candidate clips. 
\citet{DBLP:conf/iccv/HendricksWSSDR17} proposed MCN which effectively localizes language queries in videos by integrating local and global video features over time.
To improve the performance of the propose-and-rank method, some works devote to improve the quality of the proposals.
\citet{DBLP:conf/aaai/Xu0PSSS19} injected text features early on when generating clip proposals to eliminate unlikely clips and thus speed up processing and boost performance.
\citet{DBLP:conf/cvpr/ZhangDWWD19} proposed to explicitly model moment-wise temporal relations as a structured graph and devised an iterative graph adjustment network to jointly learn the
best structure in an end-to-end manner.
The others mainly worked on designing a more effective multi-modal interaction network. 
\citet{DBLP:conf/mm/LiuWN0CC18} utilized a language-temporal
attention network to learn the word attention based on
the temporal context information in the video.
\citet{DBLP:conf/sigir/LiuWN0CC18} designed a memory attention model to dynamically compute the visual attention over the query and its context
information. However, these models are sensitive to the heuristic rules.

Proposal-free approaches~\cite{DBLP:conf/aaai/YuanM019,LuCTLX19,DBLP:conf/aaai/ChenLTXZTL20,ChenCMJC18,DBLP:conf/acl/ZhangSJZ20} directly predict the probabilities for each frame whether the
frame is the boundary frame of the ground-truth video segment or regress the location. 
\citet{DBLP:conf/aaai/YuanM019} directly regressed the temporal
coordinates from the global attention outputs. 
\citet{DBLP:conf/acl/ZhangSJZ20} regarded the NLVL task as a span-based QA problem by treating the input video as a text passage and directly classified the start and end points.
In order to further improve the performance, some works focus on eliminating the problem of imbalance of the positive and negative samples. 
\citet{LuCTLX19} and \citet{DBLP:conf/aaai/ChenLTXZTL20} regarded all frames falling in the ground
truth segment as foreground, and each foreground frame regresses the unique distances from its location to bi-directional boundaries.

There are also some other works~\cite{DBLP:conf/aaai/HeZHLLW19,DBLP:conf/cvpr/WangHW19} solving the NLVL task by RL, which formulates the selection of start and end timestamps as a sequential decision making process. And some concurrent NLVL works also borrow the design of Visual Transformers~\cite{dosovitskiy2020image,wang2021crossformer} to explore Transformer-based NLVL model~\cite{cao2021on}.

\noindent\textbf{End-to-End Object Detection.}
The development of NLVL is inspired by the success of object detection methods. Object detection aims to obtain a tight bounding box and a class label for each object. It can be categorized into anchor-based and anchor-free approaches. Traditional anchor-based models~\cite{DBLP:conf/nips/RenHGS15,DBLP:conf/nips/DaiLHS16} 
have dominated this area for many years, which place a series of anchor boxes uniformly and do the classification and regression to determine the position and class for the objects. Anchor-free models~\cite{DBLP:conf/eccv/LawD18,DBLP:conf/iccv/DuanBXQH019} are becoming prosperous, which have been promoted by the development of key point detection. They directly predict key points and group them together to determine objects. Recently, end-to-end object detectors based on sparse candidates have drawn large amount of attentions. DETR~\cite{DBLP:conf/eccv/CarionMSUKZ20} utilizes a sparse set of object queries to interact with the global feature. Benefit from the bipartite matching predictions and ground-truths, DETR can discard the NMS procedure while achieving remarkable performance. Similarly, Sparse R-CNN~\cite{DBLP:journals/cvpr/Sun} provides a fixed set of sparse learnable object proposals and performs classification and localization. Our model borrows the similar idea from Sparse R-CNN.

\section{Approach}
We formally define NLVL task as follows. Given an untrimmed video as $V=\{f_t\}^T_{t=1}$ and a language query 
as $Q=\{w_n\}^M_{m=1}$, where $T$ and $M$ are the number of video frames and query words, NLVL needs to predict the 
start and end timestamps $(t_s, t_e)$ of the video segment described by the language query $Q$.  
For each video, we extract its visual features $\bm{V} = \{\bm{v}_t\}^T_{t=1}$ by a pre-trained 3D ConvNet. For each query, we initialize the word features $\bm{Q} = \{\bm{w}_n\}^M_{m=1}$ using the GloVe embeddings~\cite{DBLP:conf/emnlp/PenningtonSM14}.

\begin{figure}
  \centering
  \includegraphics[width=0.49\textwidth]{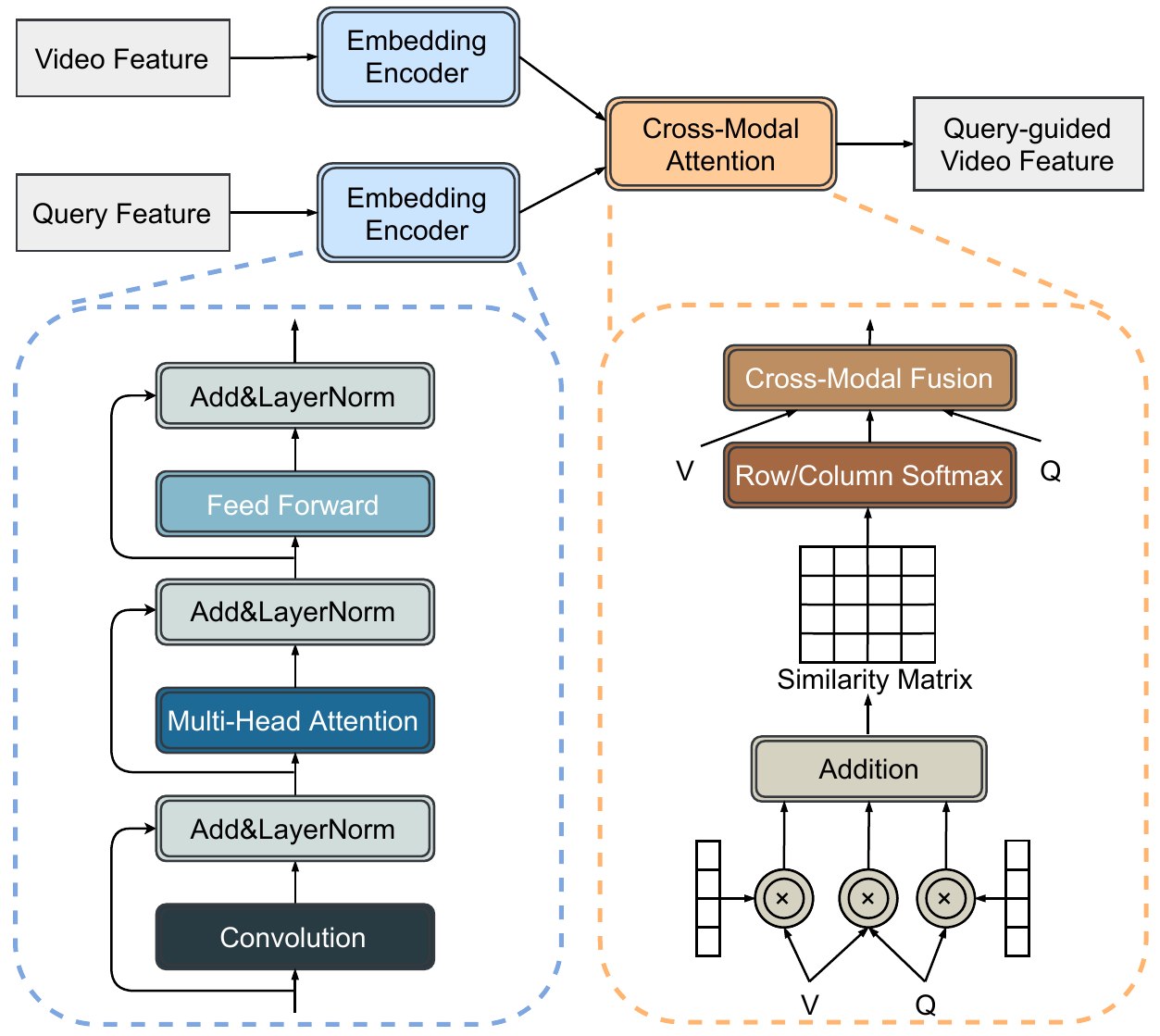}
  \caption{The overview of the Feature Encoder. The embedding encoder consists of several conv-layers, a self-attention layer and a feed-forward layer together with layer normalization and residual connection. The cross-modal attention layer learns the multi-modal interaction and outputs the query-guided video feature.}
  \label{figure3}
\end{figure}

The overall architecture of the proposed LPNet is shown in Figure~\ref{figure2}. In this section, we first introduce each component of LPNet in sequence. Then, we show the training and inference stage in details. 

\subsection{Feature Encoder}
\noindent\textbf{Embedding Encoder Layer.}
Following previous works~\cite{LuCTLX19}, we use a similar feature encoder in QANet~\cite{DBLP:conf/iclr/YuDLZ00L18}.
The embedding encoder layer consists of multiple components as shown in the left of Figure~\ref{figure3}.
The input of this layer is visual features $\bm{V} \in \mathbb{R}^{T\times d_{v}}$ and text features $\bm{Q}\in\mathbb{R}^{M\times d_{q}}$. 
We project them into the same dimension and feed them into the embedding encoder layer respectively to integrate contextual information. The output of the embedding encoder layer $\bm{V}^{'} \in \mathbb{R}^{T\times d}$ and $\bm{Q}^{'}\in\mathbb{R}^{M\times d}$ is  refined visual and text features that encode the interaction under each modality.


\noindent\textbf{Cross-Modal Attention Layer.}
This layer calculates vision-to-language attention and language-to-vision attention weights, and then encodes the multi-modal feature. 
As shown in the right of Figure~\ref{figure3}, it first computes a similarity matrix $\mathcal{S}\in\mathbb{R}^{T\times M}$, where the element $\mathcal{S}_{ij}$ 
indicates the similarity between the frame $f_{i}$ and the word $w_{j}$. 
Then, the two attention weights $\bm{A}$ and $\bm{B}$ are computed:
\begin{equation}
\bm{A} = \mathcal{S}_{row}\cdot\bm{Q}^{'}, \quad \bm{B} = \mathcal{S}_{row}\cdot\mathcal{S}_{col}^{T}\cdot\bm{V}^{'},
\label{equation1}
\end{equation}
where $\mathcal{S}_{row}$ and $\mathcal{S}_{col}$ are the row and column-wise normalization of $\mathcal{S}$.
We then model the interaction between the video and the query by the cross-modal attention layer:
\begin{equation}
\bm{V}^{q}=\mathtt{FFN}([\bm{V}^{'};\bm{A};\bm{V}^{'}\odot\bm{A};\bm{V}^{'}\odot\bm{B}]),
\label{equation2}
\end{equation}
where $\odot$ is the element-wise multiplication, and $[\cdot]$ is the concatenation operation. The $\mathtt{FFN}$ denotes feed-forward layer. The output of this layer $\bm{V}^{q}$ encodes visual features with query-guided attention.

\subsection{Propose-and-Rank Module}
\noindent\textbf{Learnable Moment Proposals.}
Different from the previous manually designed temporal anchors, 
our proposal boxes are learnable during the training process.
We define the number of proposals as $N$. 
The proposal boxes are represented by 2-d parameters ranging from 0 to 1 which are randomly initialized, denoting normalized center coordinates and lengths. 
The parameters of proposal boxes $(N\times2)$ will be dynamically adjusted with the back-propagation. In order to model the implicit relation among proposals, following~\cite{DBLP:journals/cvpr/Sun}, we attach \emph{proposal features} $\bm{P}\in\mathbb{R}^{N\times d}$ to every proposal boxes. 
Simultaneously, multi-head self-attention ($\mathtt{MHSA}$) mechanism is applied to proposal features to reason about interactions among proposals: $\bm{\widetilde{P}} = \mathtt{MHSA}(\bm{P})$. 


\noindent\textbf{Interactive Rating Layer.}\label{IRL}
Given $N$ moment proposal boxes for video $V$, we capture the candidate features
$\bm{C}$ from the visual feature $\bm{V}^{q}$ in Eq.~\eqref{equation2}.
The generated video segment candidates have different lengths in the temporal dimension,  
hence we transform the candidate features into identical length using temporal RoIAlign. For $i$-th candidate feature $\bm{C}_{i}$:
\begin{equation}
  \bm{{C}^{'}}_{i} = \mathtt{RoIAlign}(\bm{C}_{i}),
  \label{equation3}
\end{equation}
where $\bm{{C}^{'}}_{i}\in\mathbb{R}^{l\times d}$.
We then interact the candidate feature with its corresponding proposal feature 
$\bm{\widetilde{P}}_{i}\in\mathbb{R}^{d}$ to encode richer information following~\cite{DBLP:journals/cvpr/Sun}:
\begin{equation}
  \bm{\widetilde{C}}_{i} = \mathtt{Flatten}(\bm{{C}^{'}}_{i} \bm{W}_p\bm{\widetilde{P}}_{i}) \bm{W}_c,
  \label{equation4}
\end{equation}
where $\bm{W}_{p}\in\mathbb{R}^{d\times d}$ and $\bm{W}_{c}\in\mathbb{R}^{ld\times d}$ are learnable weights, and $\mathtt{Flatten}$ is an operation that flattens the matrix to one-dimensional vector.
We also obtain sentence-level query feature $\bm{\widetilde{Q}}\in\mathbb{R}^d$ by weighted pooling over word-level features $\bm{Q}^{'}$. 
Then, we fuse them by concatenation followed by a feed-forward layer:
\begin{equation}
\bm{F}_i = \mathtt{FFN}([\bm{\widetilde{C}}_{i}+\bm{\widetilde{P}}_{i},\bm{\widetilde{Q}}]),
\label{equation5}
\end{equation}
where $[\cdot]$ is the concatenation operation. Taking the multi-modal interactive feature $\{\bm{F}_i \}$ as input, this layer predicts the matching score for each video segment candidate and the language query. 
This layer consists of two feed-forward layers followed by ReLU and Sigmoid activation respectively: 
\begin{equation}
\hat{s}_{i} = \mathtt{Sigmoid}(\bm{W}_{2} \mathtt{ReLU} (\bm{W}_{1}\bm{F}_i)),
\label{equation6}
\end{equation}
where $\hat{s}_{i}$ indicates the predicted matching score of the $i$-th candidates. 
We argue that the matching scores are positively associated with the temporal
IoU scores between candidates and ground-truth moment. 
Therefore, we use the IoU scores as the ground-truth labels to supervise the training process. 
As a result, the matching score rating problem turns into an IoU regression problem. 
\subsection{Dynamic Adjustor}
\label{sec3.3}
The dynamic update of our learnable proposals is performed by the dynamic adjustor. 
For $N$ candidates generated before, we can obtain $N$ scores for each candidate from the 
previous layer. 
The update strategy is that the model only adjusts the proposal with the largest score when a sample comes in, namely the most certain update.  
Through multiple iterations, the learned proposals can statistically represent the real distribution of the dataset. 
We adopt temporal IoU loss to achieve the goal:
\begin{equation}
  \mathcal{L}_{\text{IoU}} = 1 - tIoU(\hat{b_{i}}, b_{i}),
  \label{equation7}
\end{equation}
where $\hat{b_{i}}$ is the bounding box of the best-matching candidate and $b_{i}$ is the ground-truth coordinates. 

\subsection{Boundary-aware Predictor}
We adopt a boundary-aware predictor to further boost the performance. 
LPNet is still essentially a propose-and-rank approach which is hard to model the boundary information. 
By depicting the video as a series of segments, propose-and-rank methods break down the natural structure of videos thus causes sub-optimal results. 
Instead of adding additional module to explicitly incorporate boundary information, we argue that only utilizing a boundary-aware loss can significantly improve model performance. 
The boundary-aware predictor takes frame-level multi-modal feature as input. 
A bidirectional LSTM and two feed-forward layers are used to predict distribution of the start and end timestamps, \ie,  
  \begin{gather}
  \bm{H}^{s},\bm{H}^{e} = \mathtt{BiLSTM}(\bm{V}^q), \\
  \bm{L}^s = \mathtt{FFN}_{s}(\bm{H}),
  \bm{L}^e = \mathtt{FFN}_{e}(\bm{H}),
  \label{equation89}
\end{gather}
  where $\bm{H}^{s}$ and $\bm{H}^{e}$ are the hidden states of the 
  $\mathtt{LSTM}_{s}$ and $\mathtt{LSTM}_{e}$, respectively. $\bm{L}^{s}$ and $\bm{L}^{e}$ denote the logits of start and end boundaries computed by two feed-forward layer.
In order to avoid introducing the noise caused by label uncertainty, following~\cite{DBLP:conf/wacv/OpazoMSLG20}, we relax the ground-truth label near the start and end point and adopt KL divergence to fit distributions.

\subsection{Training and Inference}\label{training}
\textbf{Training Objectives.} Each training sample consists of an untrimmed video, a language query and the corresponding ground-truth video segment. 
For each segment candidate, we compute the temporal IoU between the 
candidate and the ground-truth segment as the matching score.
For each video frame with the frame-level feature, two class labels indicating whether or not the frame is the start or the end boundary are assigned. 
In this paper, we use soft label to avoid label uncertainty. 
There are two loss functions for the propose-and-rank module and boundary-aware predictor:

\noindent\textbf{Matching Regression Loss:}
\begin{equation}
\mathcal{L}_{\text{reg}} = f_{\text{MSE}}(\hat{s},s_{\text{IoU}}),
\label{equation10}
\end{equation}
where $f_{\text{MSE}}$ is a $L2$ loss function and $s_{\text{IoU}}$ is the ground-truth temporal IoU scores. 

\noindent\textbf{Boundary-aware Loss:}
\begin{equation}
\mathcal{L}_{KL} = D_{KL}(P_{s}||Y_{s}) + D_{KL}(P_{e}||Y_{e}),
\label{equation11}
\end{equation}
where the $D_{KL}$ is Kullback-Leibler divergence. 
$Y_{s}$ and $Y_{e}$ are ground-truth relaxed labels for the start and end boundaries, respectively. $P_{s}$ and $P_{e}$ are obtained from $\bm{L}^{s}$ and $\bm{L}^{e}$ via SoftMax.

Thus, the final loss is a multi-task loss combining the $\mathcal{L}_{KL}$ and $\mathcal{L}_{\text{reg}}$, \ie,
\begin{equation}
\mathcal{L} = \mathcal{L}_{KL} + \lambda \times \mathcal{L}_{\text{reg}},
\label{equation12}
\end{equation}
where $\lambda$ is a weight that balances the two losses. 
During training, we directly use the ground-truth matching score to decide which proposal to adjust by $\mathcal{L}_{\text{IoU}}$ in Eq.~\eqref{equation7}.
Specifically, the $\mathcal{L}_{\text{IoU}}$ is used to update the parameters of proposal boxes and $\mathcal{L}$ is used to update the rest of network.


\noindent\textbf{Inference.} Given a video, a language query and a set of learned proposal boxes, we forward them through the network and obtain 
$N$ segment candidates 
with their corresponding matching scores $\hat{s}$. Then, we rank the 
$\hat{s}$ and select the candidate with the highest score as the final 
result. 

\section{Experiments}
\subsection{Datasets}\label{Dataset}
We evaluate our LPNet on two public benchmarks: 1) \textbf{Charades-STA}~\cite{DBLP:conf/iccv/GaoSYN17}: It is built on Charades and contains 6,672 videos of daily indoor activities. Charades-STA contains 16,128 sentence-moment pairs in total, where 12,408 pairs are for training and 3,720 pairs for testing. The average duration of the videos is 30.59s and the average duration of the video segments is 8.22s. 2) \textbf{ActivityNet Captions}~\cite{DBLP:conf/iccv/KrishnaHRFN17}: It contains around 20k open domain videos for video grounding task. \
We follow the split in~\cite{DBLP:conf/aaai/YuanM019}, which consists of 37,421 
sentence-moment pairs for training and 17,505 for testing.
The average duration of the videos is 117.61s and the average length of the video segments 
is 36.18s.

\subsection{Evaluation Metrics}\label{Metrics}
Following prior works~\cite{yuan2021closer}, we adopt ``R@$n$, IoU=$\theta$" and ``mIoU" as evaluation metrics. 
Specifically,``R@$n$, IoU=$\theta$" represents the percentage of the testing samples that 
have at least one of the top-N results whose IoU with the ground-truth is larger than $\theta$. ``mIoU" means the average IoU with ground-truth over all testing samples. 
In our experiments, we set $n$ = 1 and $\theta\in \{0.3, 0.5, 0.7\}$. 

\subsection{Implementation}\label{Implementation}
We down-sample frames for each video and extract visual features 
using C3D~\cite{DBLP:conf/iccv/TranBFTP15} network pretrained on Sports-1M. Then we reduce the features to 500 dimension by PCA. 
For language queries, we initialize each word with 300d GloVe vectors and all word embeddings are fixed during training.
The dimension of the intermediate layer in LPNet is set to 256. 
The number of convolution blocks in embedding encoder is 4 and the kernel size is set to 7. 
The temporal RoIAlign length $l$ is set to 16. 
To avoid elaborate design, the number of learnable proposals is uniformly set to 300 on both datasets. 
For all datasets, we trained the model for 100 epochs. 
Dropout and early stopping strategies are adopted to prevent overfitting. We implement our LPNet on Tensorflow. 
The $\lambda$ in Eq.~\eqref{equation12} is set to 100. 
The whole framework is trained by Adam optimizer with learning rate 0.0001. 

\subsection{Comparisons with the State-of-the-Arts}\label{Comparison}
\noindent\textbf{Settings.} We compare the proposed LPNet with several state-of-the-art NLVL methods on two datasets. These methods are grouped into three categories by the viewpoints of propose-and-rank and proposal-free approach: 
 1) propose-and-rank models: 
 \textbf{CTRL}~\cite{DBLP:conf/iccv/GaoSYN17},  \textbf{ROLE}~\cite{DBLP:conf/mm/LiuWN0CC18},  \textbf{ACL}~\cite{DBLP:conf/wacv/GeGCN19}, \textbf{SAP}~\cite{DBLP:conf/aaai/ChenJ19a}, 
\textbf{QSPN}~\cite{DBLP:conf/aaai/Xu0PSSS19}, 
\textbf{TGN}~\cite{ChenCMJC18},
\textbf{MAN}~\cite{DBLP:conf/cvpr/ZhangDWWD19}.
\textbf{2D-TAN}~\cite{DBLP:conf/aaai/ZhangPFL20}
2) proposal-free models:
\textbf{ABLR-af}, \textbf{ABLR-aw}~\cite{DBLP:conf/aaai/YuanM019}, \textbf{DEBUG}~\cite{LuCTLX19},
\textbf{ExCL}~\cite{DBLP:conf/naacl/GhoshAPH19},
\textbf{GDP}~\cite{DBLP:conf/aaai/ChenLTXZTL20}, \textbf{VSLNet}~\cite{DBLP:conf/acl/ZhangSJZ20}, \textbf{LGI}~\cite{DBLP:conf/cvpr/MunCH20}, \textbf{DRN}~\cite{DBLP:conf/cvpr/ZengXHCTG20}, \textbf{BPNet}~\cite{DBLP:conf/aaai/XiaoCZJSYX21}
3) Others:
\textbf{RWM}~\cite{DBLP:conf/aaai/HeZHLLW19}, \textbf{SM-RL}~\cite{DBLP:conf/cvpr/WangHW19}.

\begin{table}[htb]
  \centering
  \setlength{\tabcolsep}{0.5mm}{
  \begin{tabular}{c|c|cccc}
  \hline
  Methods  & Feat.  & IoU=0.3 & IoU=0.5 & IoU=0.7 & mIoU  \\ \hline
  CTRL     & C3D      & --      & 23.63   & 8.89    & --    \\
  ROLE     & C3D      & 25.26   & 12.12   & --      & --    \\
  ACL-K      & C3D      & --   & 30.48   & 12.20   & --    \\
  SAP      &C3D      & --      & 27.42   & 13.36   & --    \\
  RWM      &C3D      & --      & 36.70    & --      & --    \\
  SM-RL    &C3D      & --      & 24.36   & 11.17   & --    \\
  QSPN     &C3D      & 54.70    & 35.60    & 15.80    & --    \\
  DEBUG    &C3D      & 54.95   & 37.39   & 17.92   & 36.34    \\
  GDP      &C3D      & 54.54   & 39.47   & 18.49   & --     \\
  VSLNet   &C3D      & 54.38   & 28.71   & 15.11   & 37.07   \\
  BPNet     & C3D      & 55.46   & 38.25   & 20.51   & 38.03  \\
  LPNet     & C3D      & \textbf{59.14}   & \textbf{40.94}   & \textbf{21.13}   & \textbf{39.67}    
  \\ \hline
  ExCL    & I3D      & --      & 44.10    & 22.40    & --    \\
  MAN     & I3D      & --        & 46.53   & 22.72    & --       \\
  VSLNet   & I3D      & 64.30    & 47.31   & 30.19   & 45.15 \\
  BPNet    & I3D      & 65.48  & 50.75   & 31.64   & 46.34\\
  DRN    & I3D      & --  & 53.09   & 31.75   & --\\  
  LPNet  & I3D      & \textbf{66.59}  & \textbf{54.33}   & \textbf{34.03}   & \textbf{47.71} \\
  \hline
  \end{tabular}}
  \caption{Performance (\%) of ``R@$1$, IoU=$\theta$" and ``mIoU"  compared with the state-of-the-art NLVL models on Charades-STA.} 
  \label{table1}
\end{table}

\begin{table}[h]
  \centering
  \setlength{\tabcolsep}{0.25mm}{
  \begin{tabular}{c|cccc}
  \hline
  Methods  & IoU=0.3 & IoU=0.5 & IoU=0.7 & mIoU  \\ \hline
  TGN        & 43.81   & 27.93   & --      & --    \\
  QSPN      & 45.30    & 27.70    & 13.60      & --    \\
  RWM      & --      & 36.90    & --      & --    \\
  ABLR-af    & 53.65   & 34.91   & --      & 35.72 \\
  ABLR-aw     & 55.67   & 36.79   & --      & 36.99 \\
  DEBUG     & 55.91   & 39.72   & --      & 39.51 \\
  GDP        & 56.17   & 39.27   & --       & 18.49 \\
  VSLNet    & 55.17   & 38.34   & 23.52   & 40.53 \\
  DRN    & --   & 42.49   & 22.25	   & -- \\
  LGI      & 58.52   & 41.51   & 23.07	   & 41.13 \\
  BPNet      & 58.98   & 42.07   & 24.69   & 42.11 \\
  2D-TAN     & 59.45      & 44.51      & \textbf{26.54}  & --     \\
  LPNet       & \textbf{64.29}   & \textbf{45.92}   & 25.39   & \textbf{44.72} \\ \hline
  \end{tabular}}
  \caption{Performance (\%) of ``R@$1$, IoU=$\theta$" and ``mIoU" compared with the state-of-the-art NLVL models on ActivityNet Captions.} 
  \label{table2}
\end{table}

\begin{figure}[h!]
    \centering
    \includegraphics[width=0.47\textwidth]{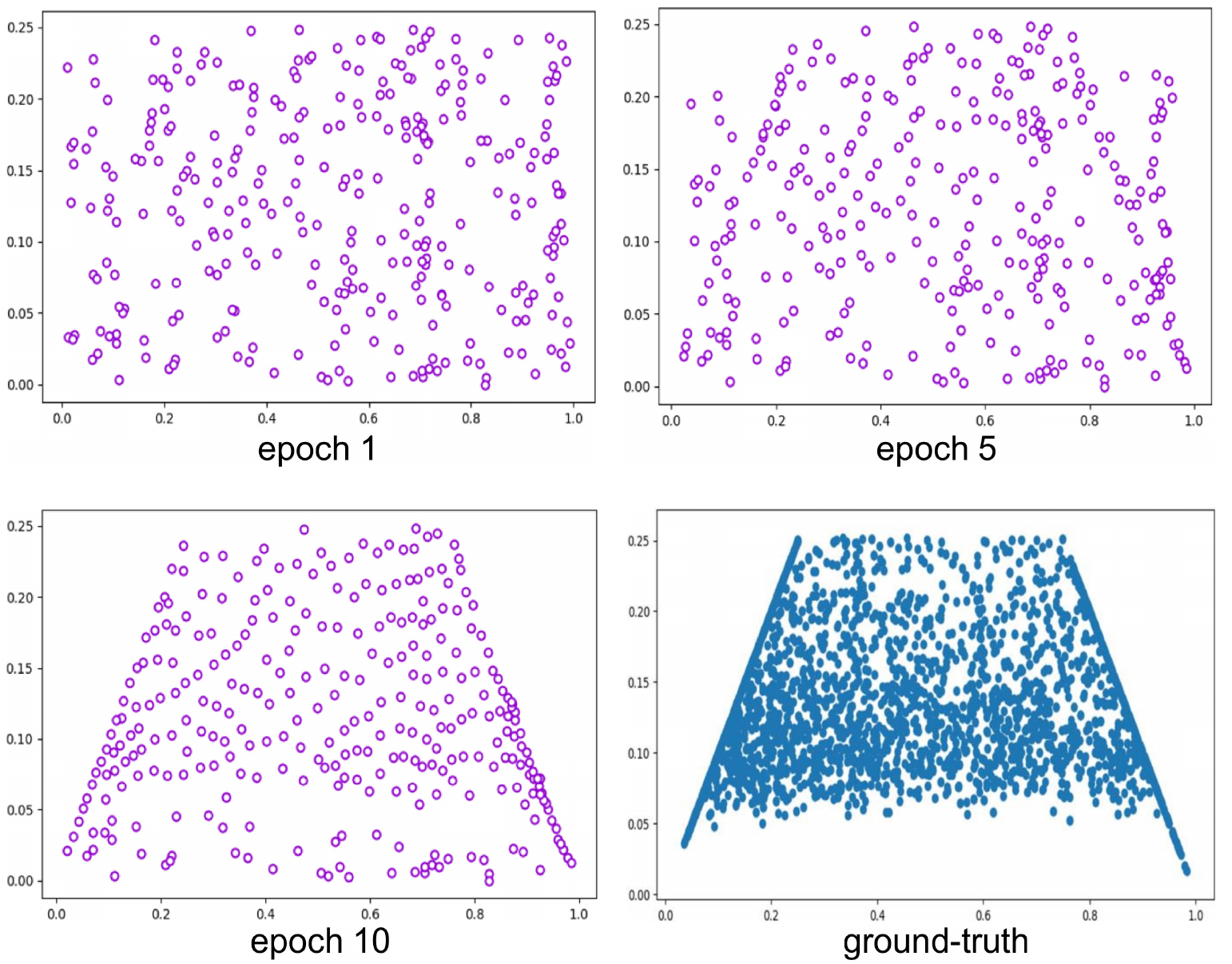}
    \caption{The distribution of learnable proposals during training process, which is getting closer to the ground-truth distribution of samples. Horizontal and vertical axes represent the normalized center coordinate and half length of proposals. We initialized the maximum length of proposals on Charades-STA as 0.5 according to its characteristics. 
    }
    \label{figure4}
\end{figure}

\begin{table}[h!]
  \centering
  \begin{tabular}{c|l|cccc}
  \hline
  \multicolumn{2}{c|}{N} &IoU=0.3 & IoU=0.5 & IoU=0.7 & mIoU\\ \hline
  \multicolumn{2}{c|}{30} & 67.45          & 53.90          & 33.52  &47.95 \\ \hline
  \multicolumn{2}{c|}{100}  & 69.09    & 55.70        & 34.97  &49.25 \\ \hline
 \multicolumn{2}{c|}{300}  & 66.59      & 54.33       & 34.03  &47.71 \\ \hline
  \end{tabular}
  \caption{Performance (\%) of LPNet with the different number on proposals on Charades-STA.}
  \label{table3}
  \end{table}
  
\begin{table}[h]
  \setlength{\tabcolsep}{0.8mm}{
    \begin{tabular}{cc|cccc}
    \hline
    MHSA & BAL & IoU=0.3 & IoU=0.5 & IoU=0.7 & mIoU \\ \hline
                &      &  65.38      & 51.05        &29.33         &45.21     \\ \hline
                \checkmark          &      &66.48        & 52.42        & 30.99        &45.87     \\ \hline
                & \checkmark    & 66.48        & 54.03        &34.22        & 47.45     \\ \hline
                \checkmark          & \checkmark    & 66.59        & 54.33       &  34.03       & 47.71     \\ \hline
    \end{tabular}}
    \caption{Performance (\%) comparisons on Charades-STA in ablative experiments of component of LPNet. MHSA: multi-head self-attention over proposal features, BAL: boundary-aware loss.}
    \label{table4}
\end{table}


The results on two benchmarks are reported in Table~\ref{table1} to Table~\ref{table2}. We can observe that our LPNet achieves new state-of-the-art performance over almost all metrics and benchmarks.
\begin{figure*}[t]
      \centering
      \includegraphics[width=0.9\textwidth]{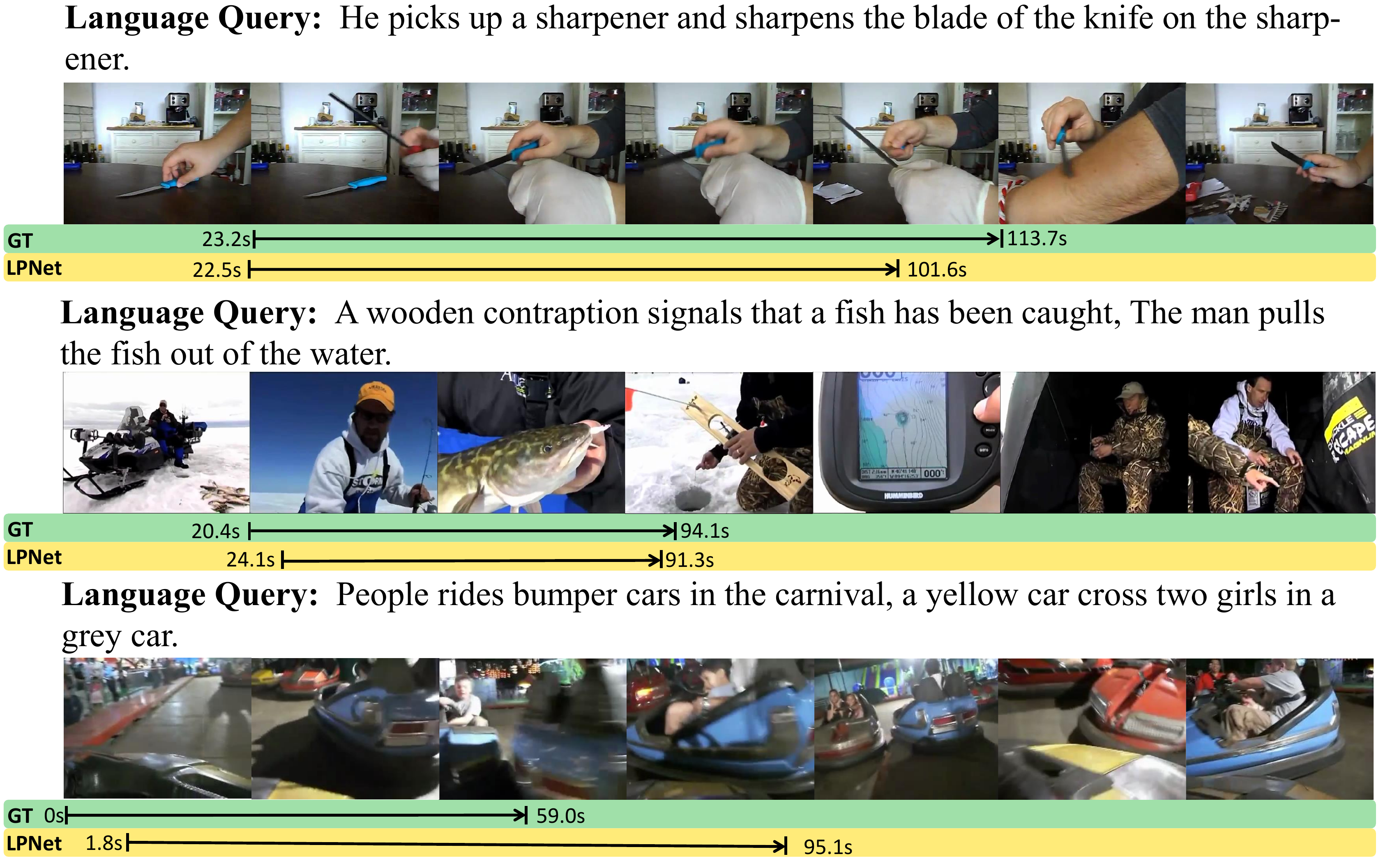}
      \caption{The qualitative results of LPNet on ActivityNet Caption dataset.}
    \label{figure5}
\end{figure*}
Table~\ref{table1} summarizes the results on Charades-STA.
For a fair comparison with methods with different feature, we use both C3D and I3D~\cite{DBLP:conf/cvpr/CarreiraZ17} video features and we report the VSLNet with C3D feature (500d by PCA) from BPNet~\cite{DBLP:conf/aaai/XiaoCZJSYX21}.
We observe that LPNet works well in even stricter metrics, 
\eg, LPNet achieves a significant 2.28 absolute improvement in $IoU@0.7$ 
compared to the second result with I3D feature, which demonstrates the effectiveness of 
our model.
To be noticed, DEBUG and VSLNet utilize the backbone similar to ours adopted from QANet.
DEBUG is a regression-based method and QANet is a classification-based method, which both belong to the proposal-free approach.
The results show that our model not only surpass a multitude of propose-and-rank methods by a lot, but also exceed these proposal-free methods. 

Table~\ref{table2} summarizes the results with C3D features on ActivityNet Captions which has longer videos in average. Our model outperforms almost all other methods. 
Compared with 2D-TAN, our LPNet achieves a significant 4.48 absolute improvement in $IoU@0.3$ but is slightly lower in $IoU@0.7$. This may be because the 2D-TAN enumerates much more candidates.
The qualitative results of LPNet are illustrated in Figure~\ref{figure4}.
We can observe that LPNet performs well to produce the precise query-related moments. 
\section{Ablation Studies}
In this section, we conduct ablative experiments with different 
variants to better investigate our approach.

\noindent\textbf{Number of Learnable Proposals.}
The number of proposals is a key factor of propose-and-rank models.
We change the number of proposals of our model on Charades-STA dataset and show
its impact on performance in Table~\ref{table3}. 
The results show that using only a small amount of proposals, LPNet is able to achieve impressive performance.
It should be noted that we simply place 300 learnable proposals on both datasets in Table~\ref{table1} and Table~\ref{table2} to avoid artificial design. 
However, a smaller amount (100) of proposals get better result on Charades-STA.

\noindent\textbf{With vs Without Boundary-aware Loss.}
From Table~\ref{table4}, we find that there is huge improvement when the boundary-aware loss is applied. 
The main reason is that the KL-divergence loss utilizes frame-level information to regularize the training of the model and force the model to consider the video as a whole. 

\noindent\textbf{With vs Without Multi-head Self-attention.}
Comparing the first two rows in Table~\ref{table4}, we observe that applying multi-head self-attention mechanism to proposal features can improve the performance. 
This operation successfully learns the latent relations between the proposals which is helpful to the localization task. However, when boundary-aware loss has been already applied (last two rows in Table~\ref{table4}), the results are very close. This may indicate that the boundary-aware loss makes the similar kind of contribution to the model.

\section{Conclusions}
In this paper, we present a novel propose-and-rank model with learnable moment proposals for NLVL. 
Compared to the existing propose-and-rank method with predefined temporal boxes, 
our model improves the performance significantly because 
1) our model disengages from the hand-designed rules for bounding boxes so that it can produce more accurate temporal boundaries; 
2) sparse sampled candidate release the pressure for subsequent ranking process; 
3) boundary-aware loss regularize the model to avoid sub-optimum. 
In the future, we are going to explore a more effective way to learn 
better proposals and extend this idea to other tasks.

\section*{Acknowledgements}
This work was supported by the National Natural Science Foundation of China (U19B2043, 61976185), Zhejiang Natural Science Foundation (LR19F020002), Zhejiang Innovation Foundation(2019R52002), and the Fundamental Research Funds for the Central Universities.
\clearpage
\bibliography{custom}
\bibliographystyle{acl_natbib}

\end{document}